\documentclass[runningheads]{llncs}
\usepackage[title]{appendix}
\usepackage[T1]{fontenc}
\usepackage[dvipsnames]{xcolor}
\usepackage{multirow}

\usepackage{graphicx}
\usepackage{makecell}
\usepackage{amsmath}
\usepackage[labelfont=bf]{caption}
\usepackage{subcaption}
\usepackage{tablefootnote}
\usepackage{algorithm2e}
\RestyleAlgo{ruled}
\usepackage{xspace}

\begin{document}

\title{DALex: Lexicase-like Selection via Diverse Aggregation\thanks{Supported by Amherst College and members of the PUSH lab.}
}

\author{Andrew Ni\inst{1}\orcidID{0000-0003-3480-4536} \and Li Ding\inst{2}\orcidID{0000-0002-1315-1196} \and 
Lee Spector\inst{3,4}\orcidID{10000-0001-5299-4797} }

\authorrunning{Ni, et al.}
\institute{Amherst College, Amherst MA 01002, USA \email{ani24@amherst.edu} \and
University of Massachusetts Amherst, Amherst, MA 01003, USA \email{liding@umass.edu}\and
Amherst College, Amherst, MA 01002, USA\and
University of Massachusetts Amherst, Amherst, MA 01003, USA \email{lspector@amherst.edu}}


\maketitle    

\begin{abstract}
Lexicase selection has been shown to provide advantages over other selection algorithms in several areas of evolutionary computation and machine learning. In its standard form, lexicase selection filters a population or other collection based on randomly ordered training cases that are considered one at a time. This iterated filtering process can be time-consuming, particularly in settings with large numbers of training cases, including many symbolic regression and deep learning applications. In this paper, we propose a new method that is nearly equivalent to lexicase selection in terms of the individuals that it selects, but which does so in significantly less time. The new method, called DALex (for Diversely Aggregated Lexicase selection), selects the best individual with respect to a randomly weighted sum of training case errors. This allows us to formulate the core computation required for selection as matrix multiplication instead of recursive loops of comparisons, which in turn allows us to take advantage of optimized and parallel algorithms designed for matrix multiplication for speedup. Furthermore, we show that we can interpolate between the behavior of lexicase selection and its ``relaxed'' variants, such as epsilon and batch lexicase selection, by adjusting a single hyperparameter, named ``particularity pressure,'' which represents the importance granted to each individual training case. Results on program synthesis, deep learning, symbolic regression, and learning classifier systems demonstrate that DALex achieves significant speedups over lexicase selection and its relaxed variants while maintaining almost identical problem-solving performance. Under a fixed computational budget, these savings free up resources that can be directed towards increasing population size or the number of generations, enabling the potential for solving more difficult problems.

\end{abstract}
\section{Introduction}
Genetic algorithms \cite{holland1992adaptation} is a subfield of evolutionary computation that uses concepts from biological evolution---random variation and fitness-based survival---to solve a wide array of difficult problems. Genetic programming (GP) is a subfield of genetic algorithms that evolves programs or functions that take some inputs and produce some outputs. Common subfields of GP include software synthesis, in which a population of programs evolves to satisfy user-defined training cases, and symbolic regression (SR), in which a population of mathematical expressions evolves to fit a regression dataset. Central to these genetic algorithms is the concept of parent selection, in which members of the current population with desirable characteristics are chosen as the starting point from which to create the next generation of individuals. Lexicase selection \cite{6920034} and epsilon lexicase selection \cite{10.1145/2908812.2908898} are state-of-the-art selection algorithms developed for the discrete-error domain and the continuous-error domain, respectively. The key idea behind lexicase selection is that it can be helpful to disaggregate the fitness function, selecting parents by considering training cases one at a time in a random order, all the while filtering out individuals which are not elite relative to the other remaining individuals on the current training case. This selection method has been shown to maintain beneficial diversity in evolved populations \cite{moore2018tiebreaks,Helmuth2016}, especially in terms of ``specialists": individuals which may have high total error but have very low errors on a subset of the training cases \cite{10.1145/3321707.3321875,Helmuth2020}. However, due to its iterative nature, lexicase selection can often take a long time, and there is not an obvious way to take advantage of parallelism or single-instruction-multiple-data (SIMD) architectures to speed up its runtime. 

In this work, we reexamine the lexicase intuition behind disaggregating the fitness function and we develop an efficient selection method based on randomly aggregating the fitness function at each individual selection event. Specifically, we quantify the idea that training cases occurring earlier in a lexicase ordering exert greater selection pressure by taking a randomly weighted average of each individual's error vector. We formulate this selection mechanism as a matrix multiplication, which allows us to take advantage of modern advancements in matrix multiplication algorithms and SIMD architectures to achieve significantly faster runtime compared to lexicase selection. 

This paper is organized as follows: In Section 2, we give a brief overview of lexicase selection and its variants. In Section 3, we describe the Diversely Aggregated Lexicase Selection (DALex) algorithm and give a brief theoretical treatment of its properties. In Section 4, we describe our experiments and results in three popular domains in which lexicase selection has seen demonstrated success. Empirical results show that DALex replicates the problem-solving success of lexicase selection and its most successful variants while offering significantly reduced runtime. 

\section{Background and Related Work}

Lexicase selection is a parent selection method that assesses individuals based on each training case in turn, instead of constructing a single scalar fitness value for each individual \cite{6920034}. During each individual selection event, the training cases are randomly shuffled. For each training case in the order determined by the random shuffle, candidate individuals in the population that are not ``elite" with respect to that training case, i.e., have an error greater than the minimum error on that training case among the remaining candidates, are filtered out. If at any point only a single individual remains, then that 
individual is selected. If all of the training cases are exhausted, then a random individual is selected from those individuals still remaining. In other words, lexicase selection chooses an individual based on the first training case, using the rest of the cases in order to break ties.
This mechanism has been shown to better maintain population diversity than methods based on aggregated fitness measures \cite{Helmuth2016}, which has been shown to improve problem-solving performance \cite{10.1145/3377929.3389987}. Lexicase selection and its variants have been successfully applied in many diverse problem domains beyond software synthesis, such as learning classifier systems \cite{10.1145/3321707.3321828}, symbolic regression \cite{10.1145/2908812.2908898}, SAT solvers \cite{metevier2019lexicase}, deep learning \cite{ding2022optimizing}, and evolutionary robotics \cite{10.1162/artl_a_00374}.  

In continuous-valued domains like SR, it is often unlikely for more than one individual to share the minimum error on a training case. Therefore, the selection pressure exerted by each training case will be too large, and lexicase selection will not proceed beyond one training case. To combat this, the epsilon-lexicase selection method was developed as a relaxation of lexicase selection \cite{10.1145/2908812.2908898}. Instead of requiring individuals to have the minimum error out of the current candidates on a training case, epsilon-lexicase selection filters out individuals with errors exceeding an epsilon threshold above the minimum error. This epsilon threshold is adaptively determined by the population dynamics, specifically as the median of the absolute deviations from the median error on each training case. 

Batch-lexicase selection \cite{10.1145/3321707.3321828,10.1145/3321707.3321793,10.1007/978-3-031-02056-8_8} is another lexicase relaxation with improved performance on noisy datasets. Instead of considering each training case in turn, batch-lexicase selection considers groups of training cases, filtering out individuals whose average accuracy on that group is lower than some threshold. Batch-lexicase selection has been successfully applied to the field of Learning Classifier Systems (LCS) in Aenugu et al. \cite{10.1145/3321707.3321828}, showing improved performance on noisy datasets compared to lexicase selection.

There have been many attempts to speed up lexicase selection. Dolson et al.~\cite{Dolson_2023} showed that the exact calculation of lexicase selection probabilities is NP-hard. However, Ding et al. \cite{10.1145/3583131.3590375} are able to calculate an approximate probability distribution from which to sample individuals. They show that their selection method achieves significantly faster runtime while achieving similar selection frequencies compared to lexicase selection. Ding et al. \cite{10.1145/3520304.3534026} also propose to use a weighted shuffle of training cases so that more difficult cases are considered first. They show that this technique can significantly  reduce the number of training cases considered while suffering minimal degradation of problem-solving power. Batch tournament selection (BTS) \cite{10.1145/3321707.3321793} orders the training cases by difficulty, then groups them into batches. BTS then performs tournament selection once per batch of cases, using the average error on that batch as the fitness function. They show that BTS achieves similar problem-solving performance to epsilon lexicase selection on SR datasets with significantly faster runtime. 

Furthermore, many selection methods based on aggregated fitness measures have been proposed for multiobjective genetic algorithms. Fitness sharing is a selection method that attempts to maintain diversity by penalizing individuals for being too close to other individuals in the population with respect to a user-defined distance metric \cite{10.5555/93126.93142}. On the other hand, implicit fitness sharing (IFS) does not require a distance metric, and instead scales the reward for each training case by the rest of the population's performance on that training case before computing the aggregated fitness function \cite{mckay2000fitness}. Historically assessed hardness (HAH) is another form of fitness sharing in which errors on each training case are scaled by the population's historical success rate on that problem \cite{klein2008genetic}. In addition to using a distance metric, NSGA-II sorts individuals into pareto fronts and conducts tournament selection using  nondomination rank and local crowding factor as fitness values \cite{deb2000fast}.

This work lies at the intersection of the lexicase variants mentioned above but takes a different direction compared to the speedup methods developed so far. For each individual selection event, we sample a random weighting of training cases. Then, for each individual, we compute a weighted average of the training case errors and select the individual with the lowest error with respect to the weighted average. In contrast to IFS or HAH, we use weights that are randomly chosen, span many more orders of magnitude, and are resampled at each selection event instead of at each generation. In contrast to NSGA-II, this method only selects non-dominated individuals because all of the training case weights are positive after the softmax operation. However, like epsilon and batch lexicase selection, and in contrast to lexicase selection, this method may select individuals outside of the ``Pareto boundary" as defined in La Cava et al. \cite{la2019probabilistic}. We formulate this selection method as a vectorized matrix multiplication, selecting all individuals in a single, batched selection event. This allows us to take advantage of modern SIMD architecture and algorithmic advances in matrix multiplication to achieve significantly faster runtime compared to lexicase selection.

\section{Diversely Aggregated Lexicase Selection}

\subsection{Description}

Diversely Aggregated Lexicase Selection (DALex) operates by selecting the individual with the lowest average error with respect to a randomly sampled set of weights. It is parameterized by the shape and scale of the distribution from which these weights are sampled. In most of our experiments, we fix the shape to that of the normal distribution and vary the scale by changing the standard deviation of the distribution. For an ablation experiment exploring the effect of different distributions on the performance of DALex, see appendix C. 
We start with a population of $n$ individuals, where each individual is evaluated on $m$ training cases. We sample an importance score for each training case from the distribution $\mathcal{N}(0,{\tt std})$ where the standard deviation {\tt std} is a tunable hyperparameter which we call the {\it particularity pressure}. Generally speaking, a training case's importance score quantifies how many times more important it is compared to the rest of the training cases. To obtain the training case weights, we softmax the importance scores. This allows us to turn differences between the values of importance scores into differences between the magnitudes of the training case weights. For each individual, we compute its average weighted error as the dot product of the training case weight vector and the individual's error vector. Finally, we select the individual with the lowest average weighted error. 

In practice, we conduct all individual selection events in one batched selection event, combining multiple weight vectors into a weight matrix and using matrix multiplication in place of the dot product. We report selection runtimes in terms of this batched selection event, i.e. the time taken to go from a population of $n$ individuals to the indices of the $n$ selected parents for the next generation.
We also perform an initial ``pre-selection'' step that differs slightly from other lexicase selection implementations due to the batched nature of our method: we group individuals into equivalence classes based on their error vectors, select equivalence classes using DALex, and then choose a random individual from each selected class \cite{helmuth2022population}. 

For a population of $n$ individuals evaluated on $m$ training cases, there will be $k\le n$ distinct equivalence classes, so the algorithm multiplies a $k\times m$ error matrix with a transposed $n\times m$ weight matrix, giving an asymptotic runtime of $O(m^2n)$, which is the same as that of lexicase selection. However, due to advances in the theory of matrix multiplication, the runtime of DALex can be reduced to $O(n^{\omega(\log_nm)})$ using methods such as the Coppersmith-Winograd algorithm \cite{coppersmith1987matrix}.

\subsection{Intuition}

For a better understanding of DALex, we provide two intuitions linking it to lexicase selection. 

    First, we argue that DALex prompts us to reconsider a central tenet of lexicase selection. The success of lexicase selection is commonly attributed to its disaggregation of the fitness function. As the reasoning goes, aggregated selection methods like tournament or fitness-proportional selection are unable to maintain effective problem solving diversity the same way lexicase selection does. To rationalize our results on DALex, we instead propose the slightly different view that the discrepancy in problem-solving performance is due to the single aggregation event preceding these selection methods, which causes a loss of information. Because lexicase selection utilizes the entire high-dimensional error vector instead of simply the average or sum, it is able to more accurately identify promising individuals. Since DALex uses many diverse aggregation events, it is also able to fully utilize the information contained in the individuals' error vectors, and is therefore able to replicate the success of lexicase selection. Even though each individual selection event operates on a single-dimensional total error, diversity arises from different random weights promoting different individuals in different selection events. 
    
     Second, we draw a correspondence between the random orderings in lexicase selection and the random weightings in DALex. Given a random ordering of training cases in lexicase selection, the  cases occurring earlier in the ordering exert more control over which individual is chosen at that selection event, i.e. have higher selection pressure. The first training case is paramount, and only if multiple individuals are equally good at the first case do we consider the second case, and so on. In DALex, the importance scores or training case weights assign explicit values to this notion of selection pressure. Given training case weights of $[1.0,~0.01,~0.1]$ for example, the first case would be ten times as important as the third case, which in turn is ten times as important as the second case. 
     In the limit of infinite particularity pressure, the differences in importance scores tend towards infinity, so the difference in magnitudes of training case weights tends to infinity, and we recover lexicase selection. As the number of training cases increases or the range of magnitudes spanned by the errors on each case increases, the particularity pressure needed to replicate lexicase selection increases. Empirically, we are able to replicate lexicase selection using modest particularity pressures such as 20. 

   This interpretation also explains why we call the standard deviation hyperparameter the particularity pressure. When importance weights are sampled with a high standard deviation, the most important training case will be much more important than the second most, and so on, just like in lexicase selection. As the standard deviation decreases, the importance of a single training case decreases. In other words, higher particularity pressures correspond to increasingly lexicase-like selection dynamics\cite{spector2023particularity}. We find that we can achieve similar successes to relaxed forms of lexicase selection such as batch-lexicase or epsilon-lexicase selection simply by choosing an appropriate particularity pressure. In this sense, DALex unifies the diverse selection methods lexicase, epsilon-lexicase, and batch-lexicase selection into a single selection method. It is important to note that DALex's only theoretical guarantee is that of asymptotically lexicase-like selection with infinite particularity pressure. As decreasing the particularity pressure is a different way of relaxation compared to those proposed for epsilon-lexicase and batch-lexicase selection, we do not expect to be able to replicate the selection dynamics of these lexicase variants to an arbitrary degree of accuracy. However, empirical results show that varying the particularity pressure gives enough flexibility to replicate the successes of these relaxed lexicase variants on the problem domains for which they were designed.

\subsection{Modifications}

For domains amenable to a relaxed lexicase variant such as epsilon-lexicase or batch-lexicase selection, we first standardize the population errors on each training case to have zero mean and unit variance. While not necessary, we find that this normalization step helps DALex perform well across many problems with a single hyperparameter setting.

Of the problem domains we study, Learning Classifier Systems (LCS) \cite{urbanowicz2009learning} has a significantly different structure and therefore requires further modification. Since an LCS individual represents a rule that matches a subset of the training cases, each individual will only have errors defined on a subset of the training cases. For these problems, we let the individual have error 0 on cases for which it is undefined. We also use the individual's support vector, which has a 1 for each training case on which the individual is defined, and a 0 otherwise. To compute the individual's average error, we take the dot product of the error vector with the weight vector and then normalize by dividing this value by the dot product of the support vector with the weight vector. 

\begin{algorithm}
\caption{Diversely Aggregated Lexicase Selection, batched selection event}\label{alg:one}
\KwData{\begin{itemize}
    \item[$\bullet$] $E=\{e_{i,j}\}$ the error value of individual $i$ on training case $j$, or 0 if individual $i$ is not defined on training case $j$
    \item[$\bullet$] $N=\{s_{i,j}\}$ the support of individual $i$ on training case $j$, which equals 1 if individual $i$ is defined on case $j$, otherwise 0
    \item[$\bullet$] $n$ the number of selection events, $m$ the number of training cases 
    \item[$\bullet$] {\tt particularity\char`_pressure}, the standard deviation of the sampled weights
    \item[$\bullet$] {\tt Relaxed}, whether to simulate a relaxed version of lexicase selection.
\end{itemize}}
\KwResult{ \begin{itemize}
\item [$\bullet$]${\tt idx}$, the $n$ indices of the selected individuals to be the parents of the next generation.
\end{itemize}
}

\If{{\tt Relaxed}}{$E$ $\leftarrow$ $\frac{E-{\tt mean}(E,{\tt axis}=0)}{{\tt std}(E, {\tt axis}=0)}$ The standardized error matrix}

$I$ $\leftarrow$ $\mathcal{N}(0,\text{\tt particularity\char`_pressure})$ The importance scores, an i.i.d gaussian matrix of size $[n,m]$

$W$ $\leftarrow$ ${\tt softmax}(I, {\tt axis}=1)$ The training case weights

$F$ $\leftarrow$ $\frac{EW^T}{SW^T}$ The normalized, weighted, average error for each individual 

${\tt idx}$ $\leftarrow$ ${\tt argmin}(F, {\tt axis=0})$ lowest error individuals w.r.t the weights $W$

\Return{ {\tt idx}}

\end{algorithm}

In problems where individuals are defined on all training cases, the support is 1 everywhere, so the normalization constant $SW^T$ is simply the sum of the training case weights, which is 1 due to the softmax operation. Under this condition, our algorithm degenerates to the simpler case considered above. The complete algorithm pseudocode, with the augmentations described above to accommodate for LCS and SR problems, is described in Algorithm 1. In our pseudocode, we define arithmetic operations as the vectorized element-wise operations used in libraries like {\tt numpy}, and batch together the $n$ importance score vectors for the $n$ individual selection events into an $n\times m$ matrix.

\section{Experiments and Results}
\subsection{CBGP}

We first compare DALex to lexicase selection on program synthesis problems, the class of problem for which lexicase selection was first developed. While we use a particularity pressure of 20 for these experiments to illustrate the robustness of DALex, we recommend setting the particularity pressure as large as possible within the range of floating point precision, or around 200, for problems suitable for lexicase selection. As DALex can exactly simulate lexicase selection in the limit of infinite particularity pressure, we also assess the ability of DALex to replicate the lexicase selection probability distribution with a moderate particularity pressure. To situate our results with respect to other lexicase approximations we also compare to plexicase selection\footnote[1]{Fixed a bug in the downsampling implementation in the released version of \cite{10.1145/3583131.3590375}}\cite{10.1145/3583131.3590375}. For plexicase selection, we use the hyperparameter setting $\alpha=1$. We use the Code-Building Genetic Programming (CBGP) system developed by Helmuth et al. \cite{pantridge2020code} and test on a suite of problems from the PSB1 Benchmark \cite{helmuth2015general} on which lexicase selection has shown good performance. For an overview on CBGP, see appendix A. We follow the default settings in CBGP, evolving 1000 individuals for 300 generations and using UMAD \cite{10.1145/3205455.3205603} with a rate of 0.09 as the sole variation operator. We present results in both the full data and downsampled paradigms. In the full data paradigm, all individuals are evaluated on all training cases. In the downsampled paradigm, a subset of the training cases is sampled at each generation and used to evaluate individuals in that generation. Downsampling has been shown to benefit the performance of Genetic Programming as we can increase the number of generations and/or individuals in the population  for a given total computational budget \cite{10.1162/artl_a_00341,10.1162/isal_a_00334,hernandez2019random}. Additional downsampling methods utilizing information from the population to choose the downsampled cases have been proposed \cite{boldi2023informed}, but we only study the randomly downsampled paradigm. In our experiments, we use a downsampling rate of 25\%. The specific problems we study are {\tt Compare String Lengths} (CSL), {\tt Median}, {\tt Number IO}, {\tt Replace Space with Newline} (RSWN), {\tt Smallest}, {\tt Vector Average}, and {\tt Negative to Zero} (NTZ).


Following the recommendation of the PSB1 benchmark, we report problem-solving performance as number of successful runs out of 100. We define a successful run as one which produces an individual with zero error on both the training cases and the unseen testing cases.
\tablefootnote[2]{Due to specific quirks of the code-building system, it is very difficult for CBGP to generalize successfully on {\tt Compare String Lengths}.}

\begin{table}
\caption{Success rates of GP runs on six benchmark problems in the full data and downsampled paradigms. The success rates across the board are very similar, and no results were significantly different between the selection methods. We perform chi-squared tests following \cite{10.1145/3583131.3590375} and underline results that were significantly worse than lexicase selection with ($p<0.05$). No results were significantly better than lexicase selection.}\label{cbgp}
\begin{tabular}{|l|c|c|c|c|c|c|}
\hline
    \multirow{2}{*}{\centering \bf Problem} & \multicolumn{3}{|c|}{\bf Success Rate}& \multicolumn{3}{|c|}{\bf Downsampled Success Rate}\\\cline{2-7}
     &{\it Lexicase}&{\it DALex}&{\it Plexicase} ($\alpha=1$)&{\it Lexicase}&{\it DALex}&{\it Plexicase} ($\alpha=1$)~\\\hline
    CSL\footnote[2]{Due to specific quirks, {\tt Compare String Lengths} is systemically difficult for CBGP to solve.} & 0& 0 &0& 0& 0 &0\\
    Median & 91 & 91 & \underline{83} & 100 & 100 & 60\\
    Number IO & 99 & 99 & 100 & 100 & 100 & 100\\
    RSWN & 12 & 15 & 6 & 66 & 68 & \underline{50}\\
    Smallest & 100 & 100 & 100 & 100 & 100 & 100\\
    Vector Average & 100 & 100 & 99& 100 & 100 & 100\\
    NTZ & 78 & 83 & 80& 99 & 100 & 100\\\hline
\end{tabular}
\end{table}

Table~\ref{cbgp} compares the success rates of DALex, plexicase, and lexicase selection on six benchmark problems under the full and downsampled paradigms. No results were significantly different between the selection methods, and DALex achieves very similar success rates to lexicase selection. For our ablation experiments on the distribution of the sampled importance scores, see appendix C. Additionally, we find that DALex runs faster than lexicase selection but slower than plexicase selection. Detailed results can be found in appendix B. 

To compare the selection dynamics of the three methods, we solve the same PSB1 problems using lexicase selection, and at every generation we sample 50,000 individuals using each selection method. From this we build up the empirical probability distributions $\{p_{it}\}$ the probability of selecting individual $i$ at generation $t$ via lexicase selection, $\{p_{it}'\}$ the probability of selecting individual $i$ at generation $t$ via DALex, and $\{p_{it}''\}$ the probability of selecting individual $i$ at generation $t$ via plexicase selection. To quantify the differences in probability distributions between lexicase selection and its approximations, we use the Jensen-Shannon divergence metric $$D_t=\frac{1}{2}\left[\sum_i p_{it}\log\left(\frac{2p_{it}}{p_{it}+q_{it}}\right)+\sum_i q_{it}\log\left(\frac{2q_{it}}{p_{it}+q_{it}}\right)\right]$$

Where $q_{it}$ is $p'_{it}$ or $p''_{it}$ for DALex and plexicase selection, respectively. The lower the JS-divergence of the DALex/plexicase selection probability distribution from the lexicase selection probability distribution, the better the approximation to lexicase selection. Furthermore, for each run in which lexicase selection produces a generalizing individual, we track that individual's lineage to find its ancestor $a_t$ in each generation $t$, of which there is only one per generation since we only use mutation operators. For each of these individuals we compute the probability ratio $r_t=\frac{q_{a_tt}}{p_{a_tt}}$ quantifying how much more likely DALex and plexicase selection are to select the successful lineage than lexicase selection. We report the average of these two metrics over all generations and all runs.

\begin{figure}
     \centering
     \begin{subfigure}[b]{0.4\textwidth}
         \centering
         \includegraphics[width=\textwidth]{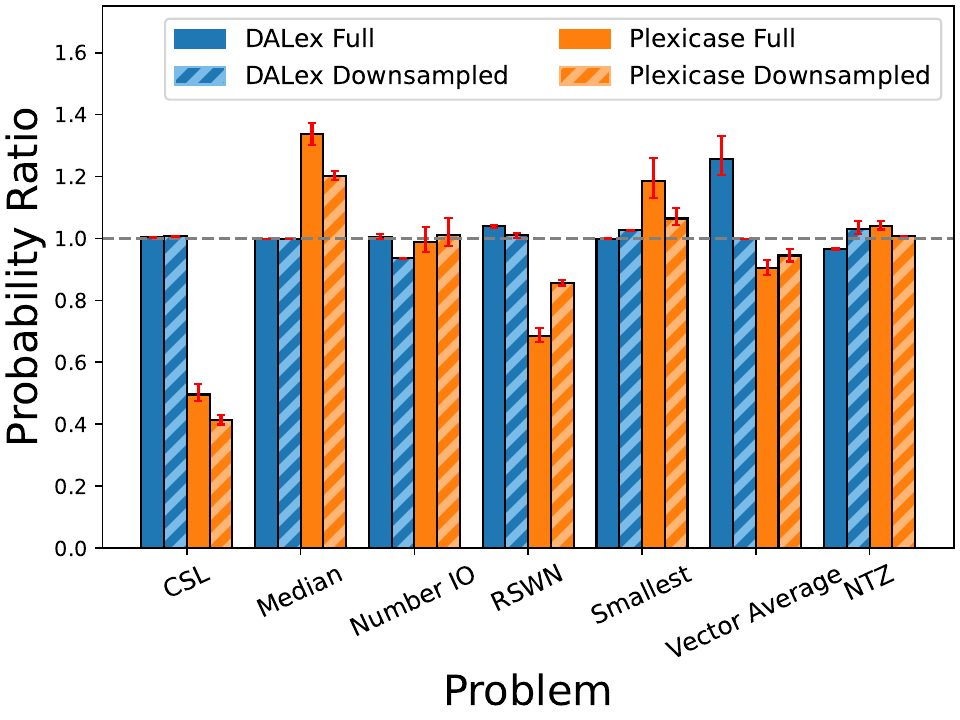}
         \caption{Probability Ratio}
         \label{fig:cbgpa}
     \end{subfigure}
     \hfill
     \begin{subfigure}[b]{0.4\textwidth}
         \centering
         \includegraphics[width=\textwidth]{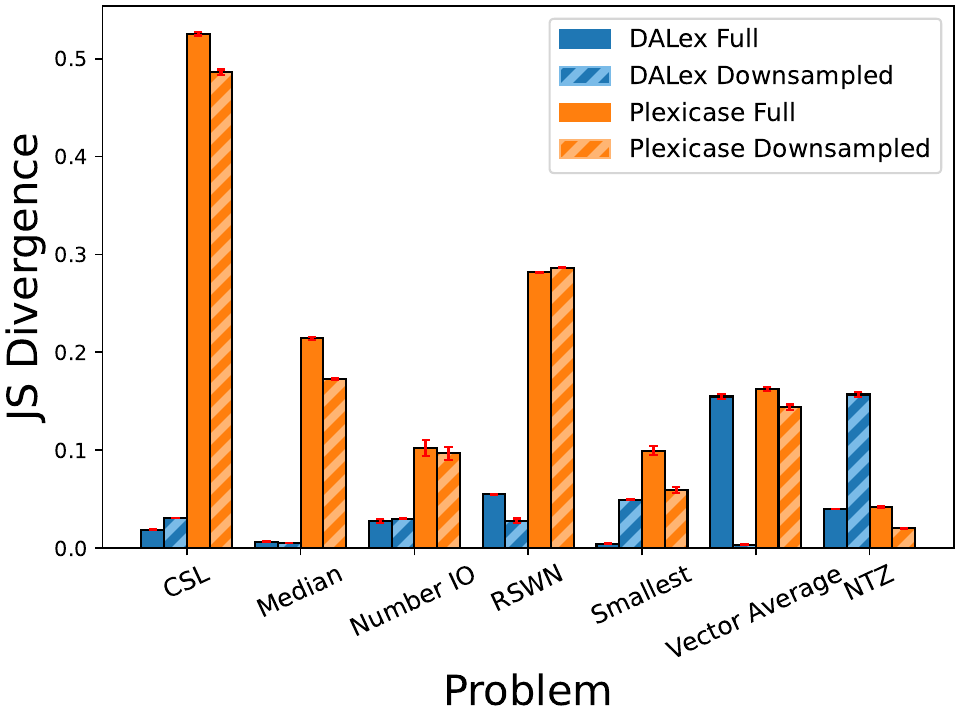}
         \caption{JS-Divergence}
         \label{fig:cbgpb}
     \end{subfigure}
        \caption{Fidelity of the DALex and plexicase approximations to lexicase selection on CBGP problems. Error bars show the bootstrapped 95\% confidence intervals. The ratio of selecting the successful lineage via DALex versus the probability using lexicase is very close to 1 in 6 out of 7 problems. The Jensen-Shannon divergence of the DALex selection probability distribution from the lexicase distribution is close to 0 in 5 out of 7 problems.}
        
        \label{fig:cbgp}
\end{figure}

Figure~\ref{fig:cbgp} shows the results for these two metrics on the problems studied. In almost all problems, the probability of selecting the successful lineage is almost the same between DALex and lexicase selection. In contrast, the probability ratio of plexicase and lexicase selection often deviates much more from 1, indicating a less faithful approximation. Furthermore, the Jensen-Shannon divergences between the DALex and lexicase selection probability distributions are often very close to zero and much smaller than the corresponding divergences of plexicase selection from lexicase selection, indicating that DALex gives a close approximation of lexicase selection dynamics. As mentioned before, we expect increasing the value of the standard deviation to improve the fidelity of the DALex approximation, bringing the DALex JS divergences even lower and the DALex probability ratios even closer to 1.

\subsection{Image Classification} 

To test the ability of DALex to mimic lexicase selection even with many training cases, we compare  DALex against lexicase selection as the selection method in gradient lexicase selection. We use the popular VGG16 \cite{simonyan2014very} and ResNet18 \cite{he2016deep} neural networks on the CIFAR10 \cite{krizhevsky2009learning} dataset. This dataset is comprised of 32x32 bit RGB images distributed evenly across 10 classes. There are 50,000 training instances and 10,000 test instances. As shown in Ding et al. \cite{10.1145/3520304.3534026}, lexicase selection often uses up to the entire training set to perform selection, especially towards the end of the genetic algorithm. 

Gradient lexicase selection\cite{ding2022optimizing} is a hybrid of lexicase selection with deep learning, in which a population of deep neural networks (DNN) is evolved to fit an image classification dataset. In gradient lexicase selection, stochastic gradient descent on a subset of the training cases is used as the mutation operator. Lexicase selection proceeds by evaluating the population of DNNs on one training case at a time and keeping only the DNNs that correctly predict the training case's label. In contrast, DALex evaluates each individual on the entire training set, assigns to each training case an error value of 0 if the case was correctly predicted or 1 otherwise, and computes a weighted sum of the error values. 
As this domain is intended to test the limits of our lexicase approximation with its large number of training cases, we use the previously recommended particularity pressure of 200. 

Since DALex is appropriate for both discrete-valued and continuous-valued errors, we also investigate the possibility of using cross-entropy loss instead of accuracy as the basis of selection. 

We use a population size of $p=4$ and run evolution for a total of $200(p+1)$ epochs. For a more detailed description of other hyperparameters such as the SGD learning rate schedule, see Ding et al. \cite{ding2022optimizing}.

Each algorithm is run 10 times, and both the final test accuracy and algorithm runtime are recorded. All runs were done on a single Nvidia A100 gpu. We report the mean and standard deviation of these two metrics in Table~\ref{vgg} the form $\text{\tt mean}\pm\text{\tt std}$. 


\begin{table}
\caption{Performance of three selection methods and two architectures on the CIFAR10 dataset. Vanilla SGD is also included as a baseline. DALex acheives similar test accuracy and lower runtime compared to lexicase selection.}\label{vgg}
\begin{tabular}{|l|c|c|c|c|}
\hline
    \multirow{2}{*}{\centering \bf Selector} & \multicolumn{2}{|c|}{\bf VGG16}& \multicolumn{2}{|c|}{\bf ResNet18}  \\\cline{2-5}
     &{\it Test Accuracy} & {{\it Runtime} (h)}&{\it Test Accuracy} & {{\it Runtime} (h)}\\\hline
     Vanilla SGD & $92.8\pm0.2$ & $0.962\pm0.003$ & $93.8\pm0.1$ &  $1.003\pm0.006$\\
     Lexicase & $93.7\pm0.1$ & $11.1\pm0.3$ & $95.1\pm0.2$ & $29\pm1$\\ 
    DALex accuracy (std=200) & $93.7\pm 0.2$ & $8.7\pm0.1$ & $95.3\pm0.1$ &$9.1\pm0.1$\\
    DALex losses (std=200) & $93.7\pm0.2$& $8.66\pm0.02$ & $95.2\pm0.2$ & $9.2\pm0.1$\\\hline
\end{tabular}
\end{table}

We find that DALex performs just as well as lexicase selection whether selecting on cross-entropy losses or accuracy, and additionally runs much faster. This is likely due to the fact that lexicase selection evaluates individuals on each training case with a batch size of 1. The small batch size causes too much communication overhead between the cpu and gpu, resulting in a slower runtime. Furthermore, as training progresses and the models approach 100\% accuracy on the training set, lexicase selection begins to exhibit its worst-case runtime. As the ResNet18 model reaches a higher training set accuracy than the VGG16 model, lexicase selection takes more than twice as long to run. With DALex, however, both models have consistent and much lower runtimes.

\subsection{SRBench}

To examine the ability of DALex to replicate epsilon lexicase-like behavior, we use the SRBench \cite{NEURIPSDATASETSANDBENCHMARKS2021_c0c7c76d} benchmark suite for symbolic regression. Specifically, we compare DALex, plexicase, and epsilon lexicase selection as the selection method for the gplearn \cite{gplearn} framework and train on 20 subsampled datasets from the Penn Machine Learning Benchmark (PMLB) \cite{romano2021pmlb} suite with sizes ranging from 50 to 40,000 instances. As recommended by SRBench, we repeat each experiment 10 times with different random seeds, and take the median of each statistic over these 10 experiments. For these experiments, we use the pre-tuned settings of 1000 individuals per generation, 500 generations, and functions drawn from the set $\{+,~ -,~ *,~ /,~ \text{log, sqrt, sin, cos}\}$. We also use the default parsimony coefficient of 0.001, which adds a model size-based penalty to the error of each test case. We choose the DALex particularity pressure to be 3, as determined from a preliminary hyperparameter search.

The statistics we examine in this paper, following the SRBench defaults, are test $R^2$, training time, and model complexity. Unlike in program synthesis, we find that the runtime of epsilon lexicase dominates the algorithm runtime, so we expect the choice of selection method to significantly impact the overall runtime. We report the median of these statistics over all 20 problems, using bootstrapping to obtain a 95\% confidence interval. 

\begin{figure}
     \centering
     \begin{subfigure}[b]{0.3\textwidth}
         \centering
         \includegraphics[width=\textwidth]{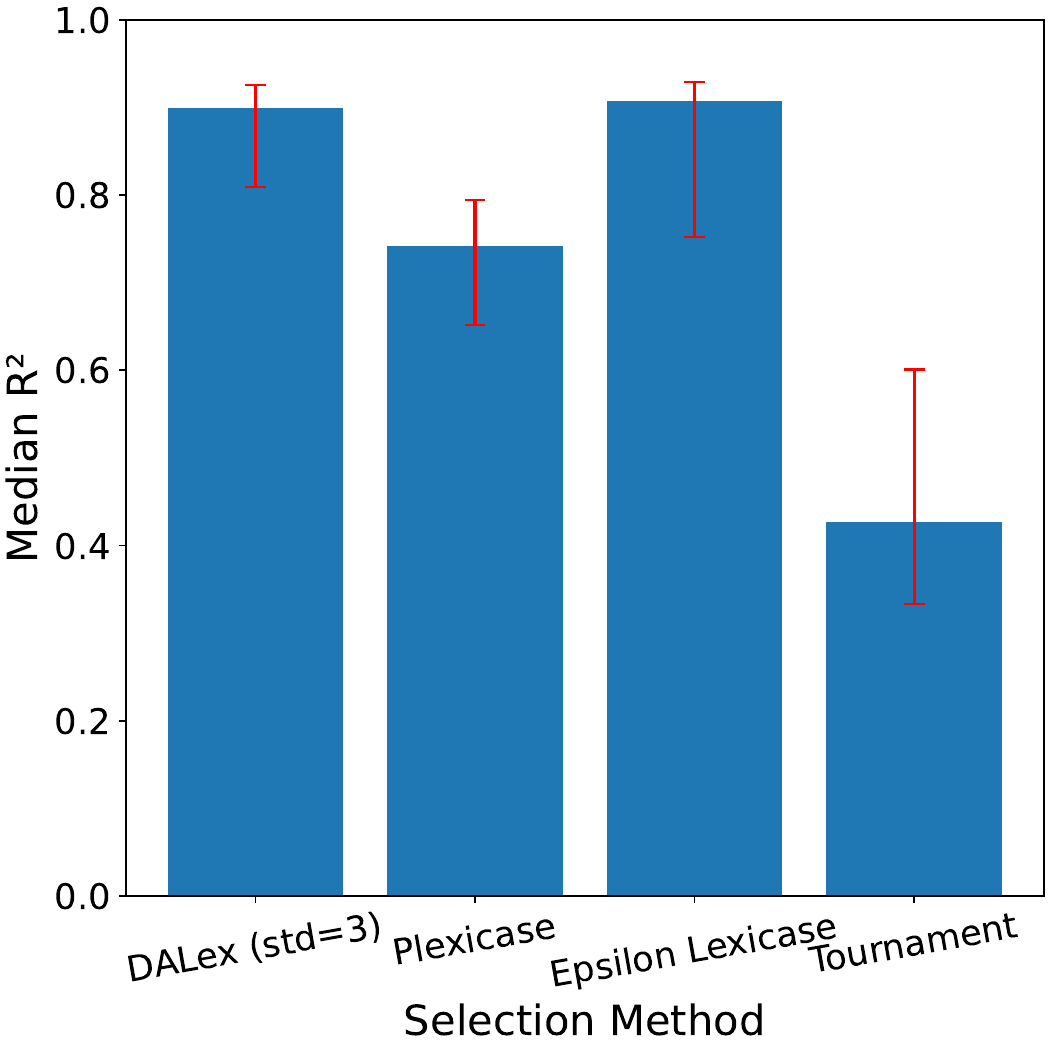}
         \caption{Test R$^2$}
         \label{fig:srbencha}
     \end{subfigure}
     \hfill
     \begin{subfigure}[b]{0.3\textwidth}
         \centering
         \includegraphics[width=\textwidth]{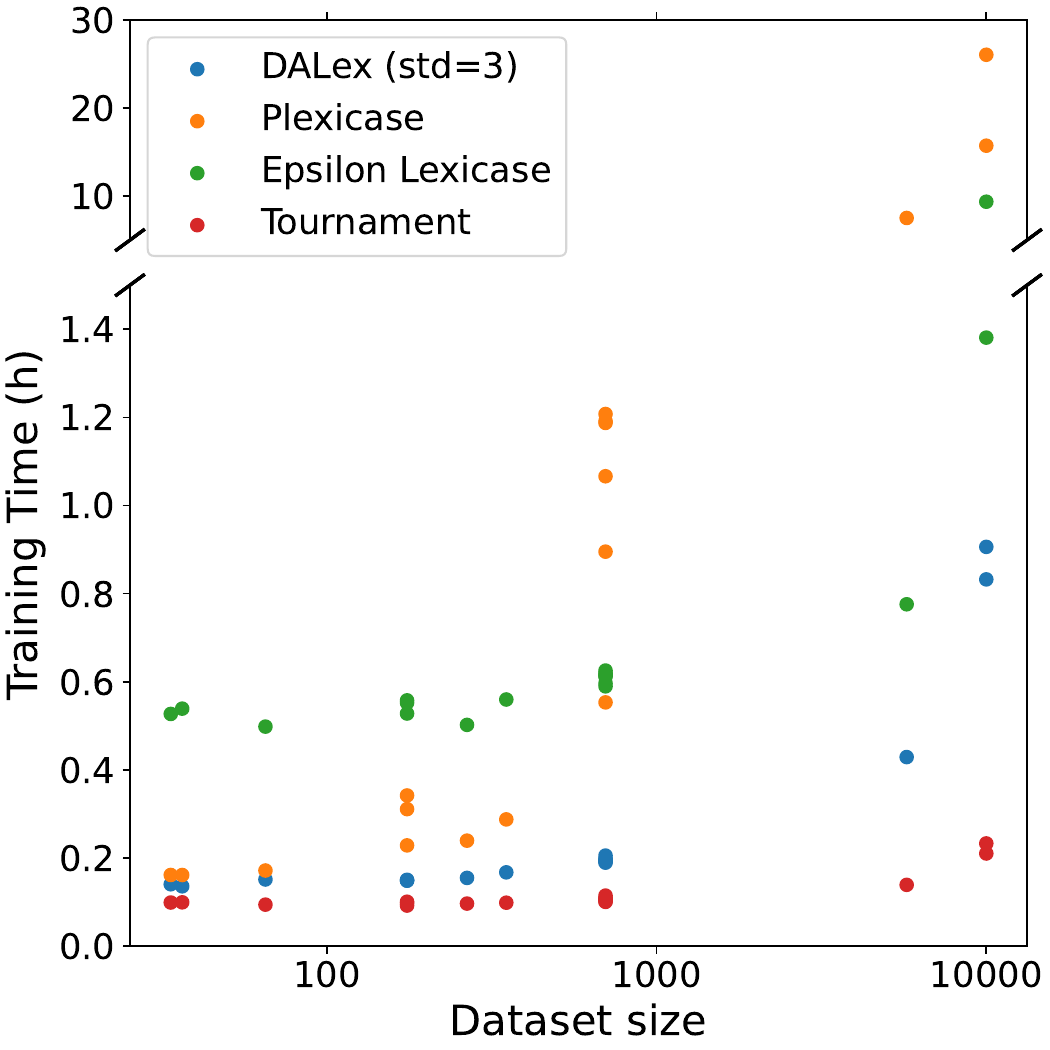}
         \caption{Runtime}
         \label{fig:srbenchb}
     \end{subfigure}
     \hfill
     \begin{subfigure}[b]{0.3\textwidth}
         \centering
         \includegraphics[width=\textwidth]{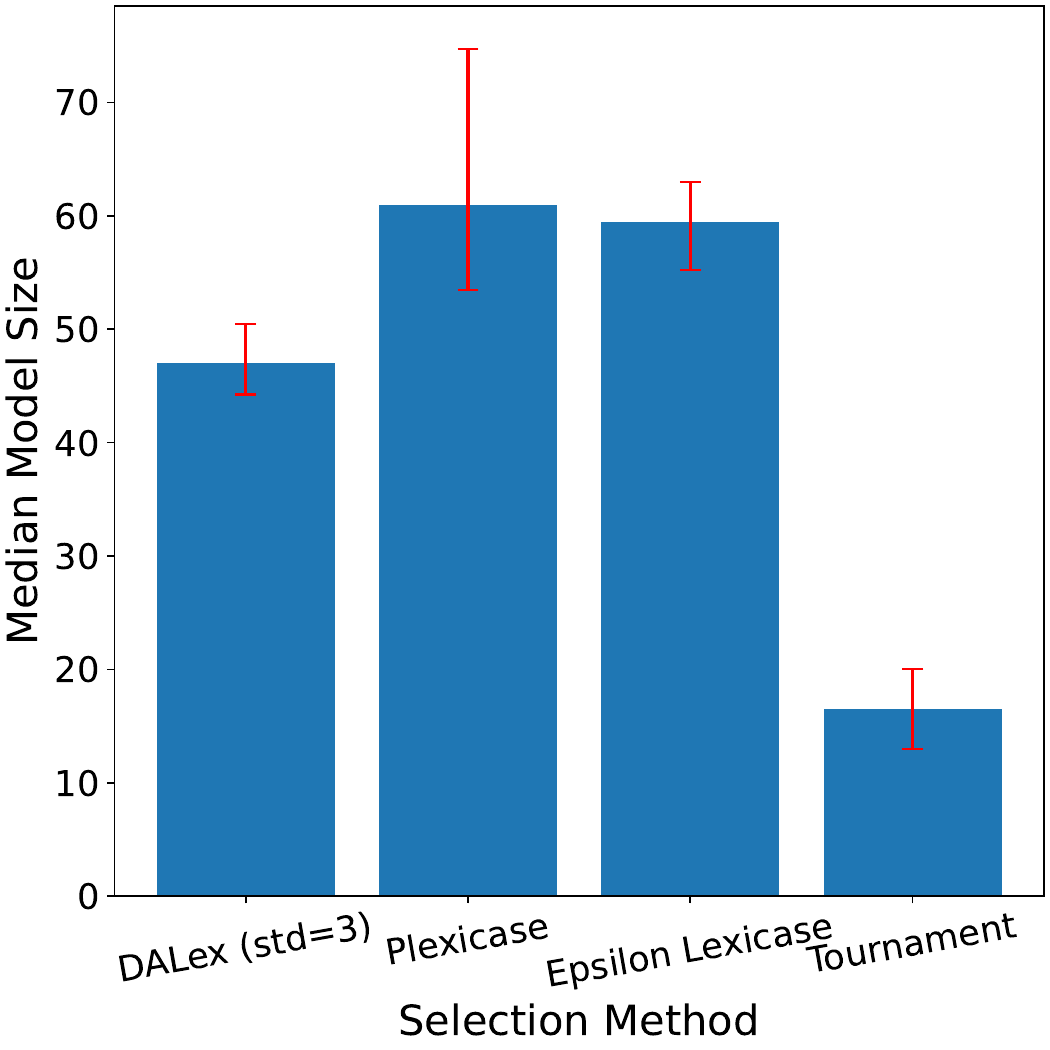}
         \caption{Model Size}
         \label{fig:srbenchc}
     \end{subfigure}
        \caption{Performance of three selection methods on 20 downsampled problems from the PMLB repository. DALex has a similar test R$^2$ to epsilon lexicase while having a lower runtime and generating more concise models.}
        \label{fig:srbench}
\end{figure}

Figure~\ref{fig:srbench} shows the results for the 20 selected problems. We find that DALex gives a very similar test R$^2$ statistic to epsilon lexicase selection while generating significantly more concise models and taking much less time to train. Of the problems we studied, epsilon lexicase selection has a much higher runtime on the {\tt 344\char`_mv} problem than on all other problems. We hypothesize that this is due to epsilon lexicase selection approaching its worst-case runtime, where it has to consider all test cases over all individuals. Helmuth et al. \cite{Helmuth2020} find that lexicase selection often selects a single individual after considering only a small fraction of the total training cases, resulting in empirical runtimes that are much faster than the worst-case runtimes. On problems like {\tt 344\_mv}, however, where the selection dynamics approach the worst-case situation, lexicase selection will have much higher runtimes than we'd expect from empirical results. In contrast, DALex has a consistent and low runtime that scales predictably with the length of individuals' error vectors. The high runtimes and degraded approximiation performance for plexicase on large datasets are new findings that call for additional investigation.

\subsection{Learning Classifier Systems}

Finally, to show that DALex can reproduce the success of batch-lexicase, we use the learning classifier system (LCS) from Aenugu et al. \cite{10.1145/3321707.3321828} We choose the led24 dataset \cite{misc_led_display_domain_57} for the stark contrast between the performances of lexicase selection and batch-lexicase selection. The led24 problem \cite{misc_led_display_domain_57}, taken from the UCI ML Repository, tests the ability of selection methods to cope with noise. This dataset, consisting of 3200 instances, contains 7 boolean attributes corresponding to whether each of 7 LED lights in a display is lit up. It has 10 categories, the digits 0 through 9, corresponding to which digit is shown on the LED display. In this dataset, each boolean attribute has a 0.1 probability having its value inverted. The evolved rule distribution for this problem is sparse, with most rules representing fewer than 5 data instances \cite{10.1145/3520304.3534026}. This property makes the led24 problem unsuitable for lexicase selection and motivates the development of batch-lexicase selection. For this domain, we use a particularity pressure of 20 as determined by a preliminary hyperparameter search.

As in Aenugu et al. \cite{10.1145/3321707.3321828}, we conduct experiments in both the full data and partial data paradigms. In the full data paradigm, we use the entire led24 dataset, leaving out 30\% for testing. In the partial data paradigm, we randomly downample the dataset, removing 40\% of the instances. We repeat this downsample for each run, meaning the instances available for each run will be different. We then split the remaining 60\% of the instances into a 40\% training set and a 20\% testing set.

\begin{figure}
     \centering
     \begin{subfigure}[T]{0.27\textwidth}
         \centering
         \includegraphics[width=\textwidth]{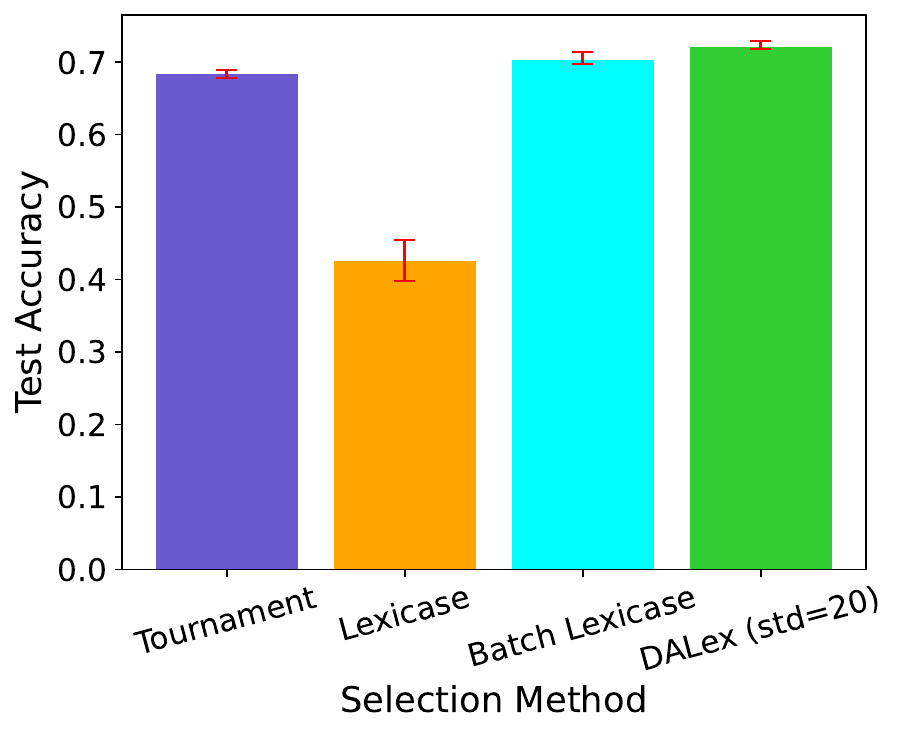}
         \caption{Accuracy (Full)}
         \label{fig:LCS1a}
     \end{subfigure}
     \hfill
     \begin{subfigure}[T]{0.27\textwidth}
         \centering
         \includegraphics[width=\textwidth]{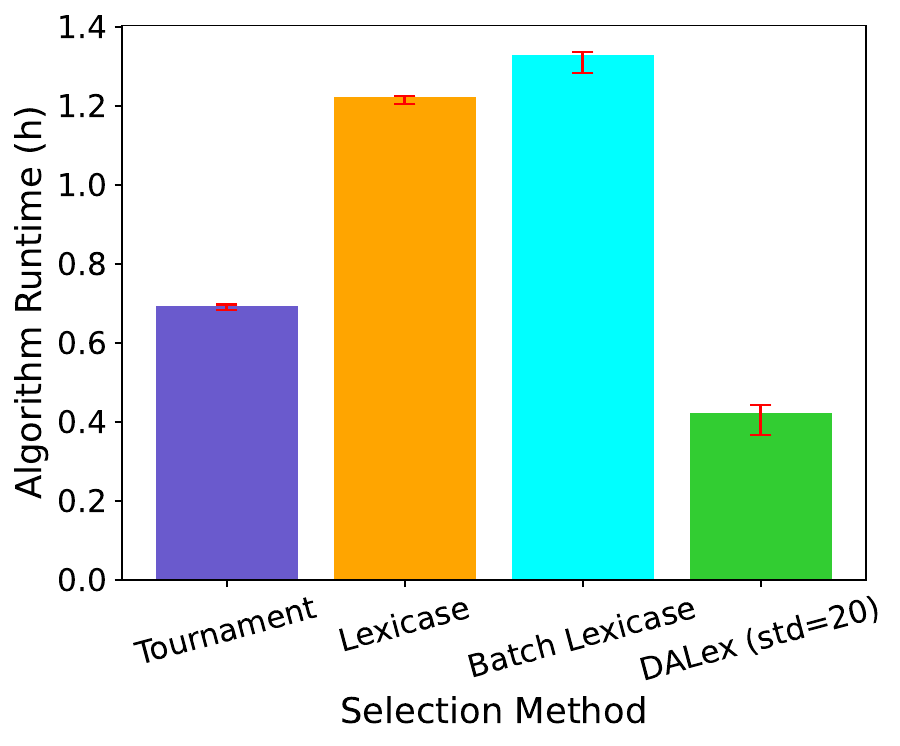}
         \caption{Runtime (Full)}
         \label{fig:LCS1b}
     \end{subfigure}
     \hfill
     \begin{subfigure}[T]{0.27\textwidth}
         \centering
         \includegraphics[width=\textwidth]{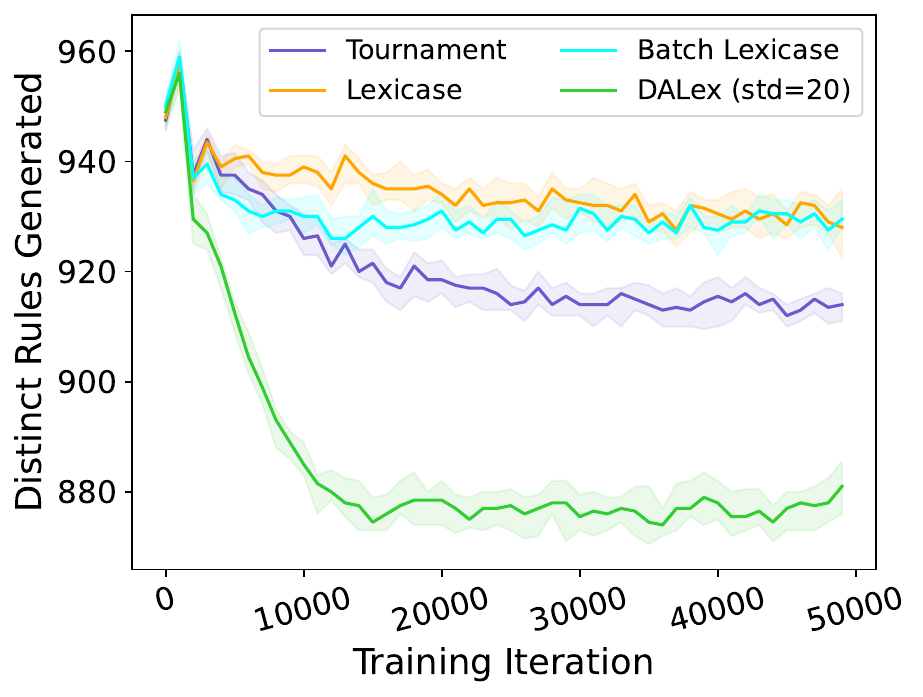}
         \caption{Rulecount (Full)}
         \label{fig:LCS1c}
     \end{subfigure}
     
     \centering
     \begin{subfigure}[T]{0.27\textwidth}
         \centering
         \includegraphics[width=\textwidth]{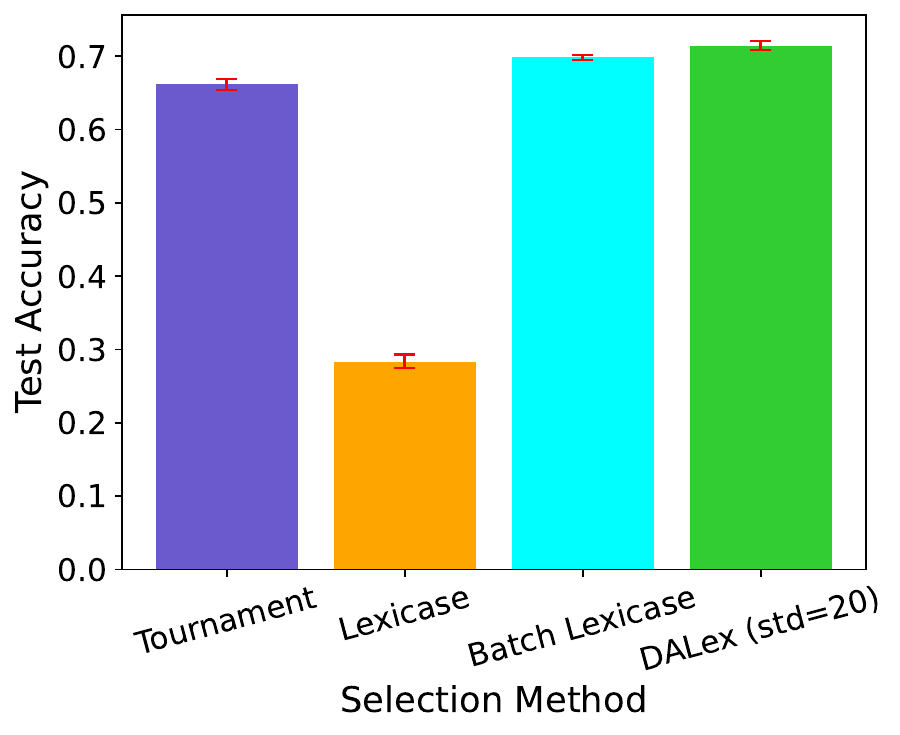}
         \caption{Accuracy (Partial)}
         \label{fig:LCS2a}
     \end{subfigure}
     \hfill
     \begin{subfigure}[T]{0.27\textwidth}
         \centering
         \includegraphics[width=\textwidth]{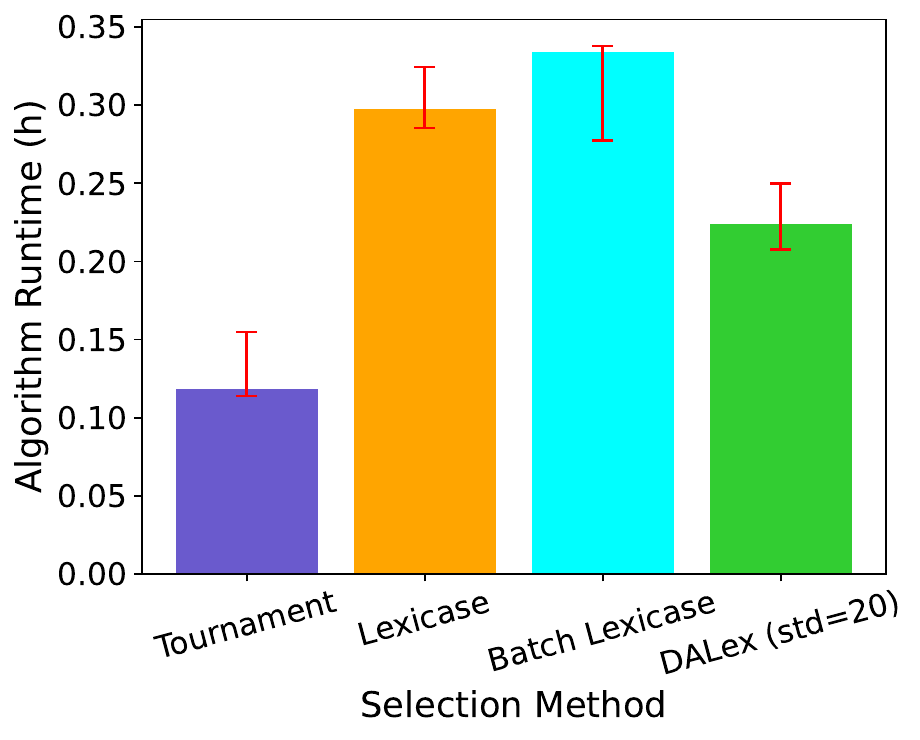}
         \caption{Runtime (Partial)}
         \label{fig:LCS2b}
     \end{subfigure}
     \hfill
     \begin{subfigure}[T]{0.27\textwidth}
         \centering
         \includegraphics[width=\textwidth]{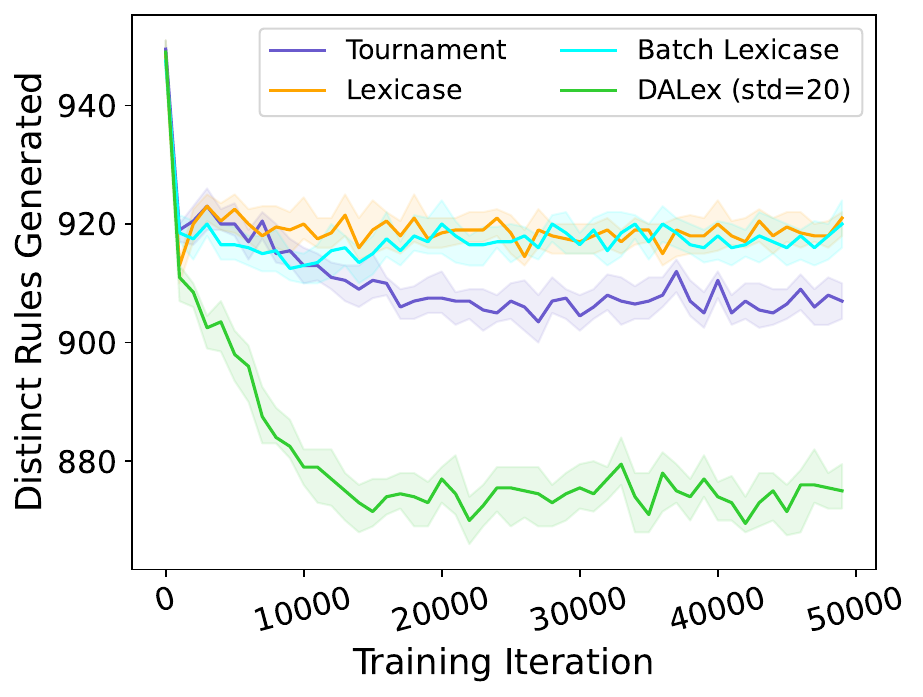}
         \caption{Rulecount (Partial)}
         \label{fig:LCS2c}

    \end{subfigure}
    
        \caption{Performance of four selection methods on the led24 problem in the full data and partial data scenarios. Bootstrapped 95\% confidence intervals are displayed as error bars for the accuracy and runtime plots and as a shaded region for the rulecount plot. DALex has approximately equal test accuracy to batch lexicase while having much lower runtime and generating the fewest rules.}
        \label{fig:LCS}
\end{figure}

Figure~\ref{fig:LCS} displays the results of our experiments. In both domains, we find that DALex has both the highest test accuracy of the four selection methods and generates the least number of distinct rules, which suggests that it may have better generalization ability. Finally, DALex has a significantly lower runtime in both domains than lexicase and batch-lexicase selection.

\section{Conclusion and Future Work}

In this work, we presented the novel parent selection method DALex, which efficiently approximates the lexicase selection mechanism by using random aggregations of error vectors. The proposed method, parametrized by the particularity pressure, is flexible enough to replicate the successes of lexicase selection and its relaxed variants on multiple problem domains, and is robust to very large datasets. Furthermore, it has a consistent and low runtime, which frees up resources that can be used for the other aspects of evolution such as number of generations. We make the code used in these experiments available at https://github.com/andrewni420/DALex.

While we have used reasonable presets for the standard deviation hyperparameter in DALex, we do not anticipate that a single hyperparameter setting will work for all cases or even for all generations during the course of a GP run. Therefore, future work could determine the optimal standard deviation using hyperparameter search or even a more radical paradigm such as autoconstruction \cite{spector2001autoconstructive,spector2002genetic},  Additionally, since DALex computes a randomly weighted fitness function, it can be combined with selection methods other than pure elitist selection, such as tournament or fitness-proportional selection \cite{goldberg1991comparative} to increase diversity. In fact, the weighted aggregation formulation of DALex is not confined to parent selection algorithms. The concept of randomly weighted aggregation could also be used to transform the losses for each training case in deep learning. By introducing more stochasticity into training, this could enable models to be trained with large batch sizes without suffering performance degradation~\cite{DBLP:journals/corr/KeskarMNST16}. Alternatively, DALex could be used to randomly weight each node in the policy head of a reinforcement learner when sampling trajectories, improving the exploration of off-policy learning algorithms such as Double Q-learning \cite{hasselt2010double} or SAC \cite{haarnoja2018soft}.

\section{Acknowledgements}

This work was performed in part using high-performance computing equipment at Amherst College obtained under National Science Foundation Grant No. 2117377. Any opinions, findings, and conclusions or recommendations expressed in this publication are those of the authors and do not necessarily reflect the views of the National Science Foundation. The authors would like to thank Ryan Boldi, Bill Tozier, Tom Helmuth, Edward Pantridge and other members of the PUSH lab for their insightful comments and suggestions.

\newpage
\bibliographystyle{splncs04}
\bibliography{refs}

\newpage 
\appendix

\section{Code-Building GP for Program Synthesis}

Code Building Genetic Programming (CBGP)\cite{pantridge2020code} is a typed genetic programming system that aims to use constructs like those found in human programming to solve program synthesis problems. Incorporating concepts like data types, data structures, and control flow into a GP system not only promises to create more complex hierarchical systems but also produces more interpretable programs which are analogous in structure to human created programs. 

CBGP encodes programs as directed acyclic graphs (DAGs) where each leaf node is an input or a constant, and each inner node a function. Together, these nodes are called ``expressions" and form a computational graph. In these graphs, the return type of child nodes are required to be sub-types of the corresponding argument in the parent node for type safety. The CBGP implementation used in this paper supports the expression types {\it Constants}, {\it Inputs},
{\it Functions}, {\it Methods}, {\it Constructors}, and {\it Higher Order Functions}. 

While CBGP programs are represented by a graph structure, it still uses the linear plushy genome representation found in other GP systems like PushGP. This representation consists of a sequence of expression tokens representing program syntax and structure tokens representing the hierarchical structure of the program. These plushy genomes are compiled into Push programs which are then translated into CBGP DAGs for execution. As in PushGP, CBGP uses the Uniform Mutation by Addition and Deletion (UMAD) mutation operator\cite{10.1145/3205455.3205603} as the sole variation operator.

\section{Program Synthesis Runtime Comparisons}

\begin{figure}
     \centering
     \begin{subfigure}[b]{0.45\textwidth}
         \centering
         \includegraphics[width=\textwidth]{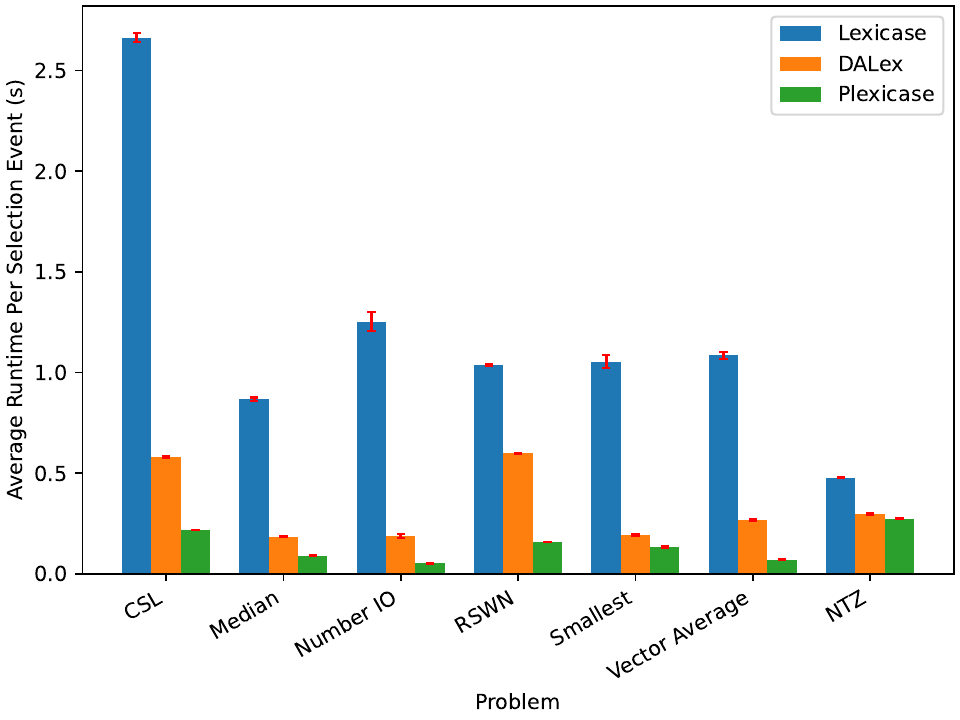}
         \label{fig:cbgpa}
     \end{subfigure}
        \caption{Runtime comparison of three selecion methods on program synthesis problems. 95\% confidence intervals obtained by bootstrapping. }
        
        \label{fig:cbgp-time}
\end{figure}

During our program synthesis experiments, we record the time taken to select the parents of the next generation. We then take the average of this metric over all generations and all runs and report them in Figure~\ref{fig:cbgp-time}. The results show that DALex is up to 5x faster than lexicase selection. Despite having a similar or slower runtime compared to plexicase selection, DALex is a more faithful approximation to lexicase selection and has shown better performance as presented in Table~\ref{cbgp}, where plexicase selection sometimes produces worse results than lexicase selection.

\section{Ablation of DALex probability distributions}

As an ablation experiment, we examine the effect of the distribution from which importance scores are chosen on the problem-solving ability of DALex. We compare our default selection from a normal distribution to selection from a uniform distribution and shuffling a range of evenly spaced numbers. The range case is interesting in that it is guaranteed to produce lexicase selection when the spacing between weights is larger than the magnitude difference between the largest possible error and the smallest possible nonzero error. In each case the importance scores are scaled so that they have a standard deviation of 20 prior to softmaxing. 

\begin{table}
\caption{Success rates of GP runs using different distributions for DALex. All distributions were scaled to a standard deviation of 20 unless otherwise noted. We perform chi-squared tests following \cite{10.1145/3583131.3590375} and underline results that were significantly worse than the normal distribution with ($p<0.05$). No results were significantly better than the normal distribution.}\label{cbgp-dist}
\centering
\begin{tabular}{|l|c|c|c|}
\hline
    \multirow{2}{*}{\centering \bf Problem} & \multicolumn{3}{|c|}{\bf Success Rate}  \\\cline{2-4}
     &~{\it Normal}~&~{\it Uniform}~&~{\it Range}~\\\hline
    CSL & 0& 0 & 0\\
    Median &  91 & 88 &87\\
    Number IO &  99 & 99 & 100\\
    RSWN&  15 &15 & 18\\
    Smallest &  100 & 100 & 100\\
    Vector Average &  100 & 100 & 100\\
    NTZ& 83 & \underline{65} & \underline{66}\\
    NTZ (std=100) & 79 & 85 & 72\\\hline
\end{tabular}
\end{table}

The results are shown in Table \ref{cbgp-dist}. The different distribution choices perform almost identically on the problems considered except for the {\tt Negative To Zero} problem. We hypothesize that this problem is more taxing for our lexicase approximations, and highlights the differences between the distributions used for DALex. Of the three distributions, the normal distribution has the largest kurtosis. Therefore, the largest importance scores sampled from the normal distribution will be farther apart than for the other two distributions. Since lexicase selection often selects an individual after considering only a fraction of the training cases \cite{Helmuth2020}, it is these largest importance scores which determine the success of DALex. The uniform distribution and random shuffle, on the other hand, lower kurtosis, and perform worse compared to the normal distribution. However, increasing the standard deviation recovers the success rate of the normal distribution.

\end{document}